\pdfoutput=1

\documentclass[11pt]{article}

\usepackage{acl}

\usepackage{times}
\usepackage{latexsym}

\usepackage[T1]{fontenc}

\usepackage[utf8]{inputenc}

\usepackage{microtype}

\usepackage{inconsolata}

\usepackage{graphicx}

\usepackage{caption}
\usepackage{listings}
\usepackage{pgfplots}
\usepackage{pgfplotstable}
\pgfplotsset{compat=1.17}
\usepackage{tikz}

\usepackage[normalem]{ulem}
\usepackage{algorithm}

\usepackage{multirow}
\usepackage{amsmath} 
\usepackage{enumitem}

\usepackage{amsmath}
\usepackage{amssymb}
\usepackage{booktabs}
\usepackage{authblk}

\usepackage{algorithm}
\usepackage{algpseudocode}
\usepackage{algorithm}
\usepackage{algpseudocode}

\usepackage[most]{tcolorbox}

\newcommand*{\affaddr}[1]{#1}
\newcommand*{\affmark}[1][*]{\textsuperscript{#1}}

\title{Prompt-Guided Internal States for Hallucination Detection of Large Language Models}

\author{
  Fujie Zhang\affmark[\S], 
  Peiqi Yu\affmark[\S],
  Biao Yi\affmark[\ddag]\thanks{\ Corresponding author.},
  Baolei Zhang\affmark[\ddag],
  Tong Li\affmark[\ddag],
  Zheli Liu\affmark[\ddag] \\
  \affaddr{\affmark[\S] School of Mathematical Sciences, Nankai University}\\
  \affaddr{\affmark[\ddag] College of Computer Science, Nankai University} \\
    \texttt{\{fujie.zhang, peiqi.yu, yibiao, zhangbaolei\}@mail.nankai.edu.cn}\quad 
    \\
    \texttt{\{tongli, liuzheli\}@nankai.edu.cn}  \\}

\begin{document}
\maketitle
\begin{abstract}
Large Language Models (LLMs) have demonstrated remarkable capabilities across a variety of tasks in different domains.
However, they sometimes generate responses that are logically coherent but factually incorrect or misleading, which is known as LLM hallucinations. Data-driven supervised methods train hallucination detectors by leveraging the internal states of LLMs, but detectors trained on specific domains often struggle to generalize well to other domains. In this paper, we aim to enhance the cross-domain performance of supervised detectors with only in-domain data. We propose a novel framework, prompt-guided internal states for hallucination detection of LLMs, namely PRISM.
By utilizing appropriate prompts to guide changes to the structure related to text truthfulness in LLMs' internal states, we make this structure more salient and consistent across texts from different domains. We integrated our framework with existing hallucination detection methods and conducted experiments on datasets from different domains. The experimental results indicate that our framework significantly enhances the cross-domain generalization of existing hallucination detection methods\footnote{We have open-sourced all the code and data in GitHub: \url{https://github.com/fujie-math/PRISM}}.
%Our code is available at \url{https://anonymous.4open.science/r/PRISM-7E2B}.

\end{abstract}

\section{Introduction}

In recent years, Large Language Models (LLMs) have demonstrated remarkable capabilities across a variety of tasks in different domains~\cite{dinan2018wizard, brown2020language, zhang2022opt, chowdhery2023palm,touvron2023llama1}. However, the hallucination problem in LLMs poses a potential threat to their practical application in many scenarios. LLM hallucinations refer to the cases where LLMs generate responses that are logically coherent but factually incorrect or misleading~\cite{zhou2020detecting, ji2023survey,DBLP:journals/corr/abs-2403-06448}. These hallucinated responses may be blindly accepted, leading users to learn incorrect information or take inappropriate actions. Therefore, detecting hallucinations in LLM-generated content becomes particularly meaningful. By using hallucination detectors to help identify incorrect generated content, users can be alerted to verify the accuracy of LLMs' responses, thus preventing potential issues.

The unsupervised paradigm focuses on assessing the confidence of LLM-generated content and rejects the low-confidence outputs.
For example, the probability information of each token generated by LLMs can serve as a measure of hallucination~\cite{kadavath2022language,zhang2023enhancing,quevedo2024detecting,hou2024probabilistic}. Additionally, the consistency among multiple responses generated by LLMs to the same question can also be used to evaluate their confidence~\cite{manakul2023selfcheckgpt}. Furthermore, \citet{chen2024inside} shifts the consistency judgment from multiple responses to their corresponding activation values in LLMs. However, these methods often struggle to achieve ideal detection accuracy or require a significant amount of additional response time~\cite{DBLP:journals/corr/abs-2403-06448}. 

For this reason, researchers have begun exploring the use of data-driven supervised methods for hallucination detection. These methods are generally based on the assumption that LLMs can recognize they have generated incorrect content, which is reflected in specific patterns in their internal states. \citet{marks2023geometry} reveal that true and false statements have discernible geometric structures in LLMs' internal states, allowing us to build classifiers by learning this structure. \citet{marks2023geometry} and \citet{sky2024androids} utilize linear structures to distinguish between different statements, while \citet{azaria2023internal} train neural networks to act as hallucination detectors.

Supervised detectors trained on specific domains often struggle to achieve good generalization performance in other domains. For example, \citet{burger2024truth} found that when distinguishing between true and false statements, the structural differences in LLMs' internal states are significantly different for affirmative and negated sentences. To address this issue, many studies focus on constructing diverse datasets or performing feature selection within LLMs' internal states to achieve better generalization performance~\cite{chen2023hallucination,burger2024truth,liu2024universal}. However, these methods require collecting additional training data from other domains, which is resource-intensive. In this paper, we aim to answer the following question:

\begin{quote}
\vspace{-0.2cm}
\textbf{\textit{Can we enhance the cross-domain performance of supervised detectors with only in-domain data?}} 
\vspace{-0.2cm}
\end{quote}

Driven by this research question, we propose a novel framework called PRISM, which stands for prompt-guided internal states for hallucination detection of LLMs. By utilizing appropriate prompts to guide changes to the structure related to text truthfulness in LLMs' internal states, we make this structure more salient and consistent across texts from different domains, which enables detectors trained in one domain to also perform well in others without additional data. Our approach is based on the insight that while LLMs' internal states encode rich semantic information, they are primarily optimized during pre-training to predict the next token rather than for hallucination detection. As a result, the directly extracted internal states contain lots of domain-specific information that is not related to text truthfulness, leading to detectors that are specific to certain domains and unable to achieve good cross-domain generalization performance.

Due to the powerful ability of LLMs to understand and follow instructions, we explore the use of prompt-guided methods to generate internal states more focused on text truthfulness. We employed various methods to investigate the effect of prompts from different perspectives. The findings indicate that the introduction of appropriate prompts can significantly improve the salience of the structure related to text truthfulness in LLMs' internal states and make this structure more consistent across different domain datasets. We also provide a simple and effective method to generate and select appropriate prompts for hallucination detection tasks. By combining prompt templates with the text to be evaluated and inputting them into LLMs, we can obtain internal states that are better suited as features for hallucination detection tasks. Subsequently, we can integrate the prompt-guided internal states with existing hallucination detection methods to construct more advanced detectors.

In summary, the contributions of this paper are as follows:
(1) We conduct an in-depth investigation into how specific prompts guide the change to the structure related to text truthfulness in LLMs' internal states.
(2) We propose PRISM, a novel framework that utilizes prompt-guided internal states to enhance hallucination detection of LLMs. 
(3) We integrate PRISM with existing hallucination detection methodologies, demonstrating its ability to significantly enhance 
the generalization performance of detectors across different domains.

\begin{figure*}[t]
\centering
    \includegraphics[width=\textwidth]{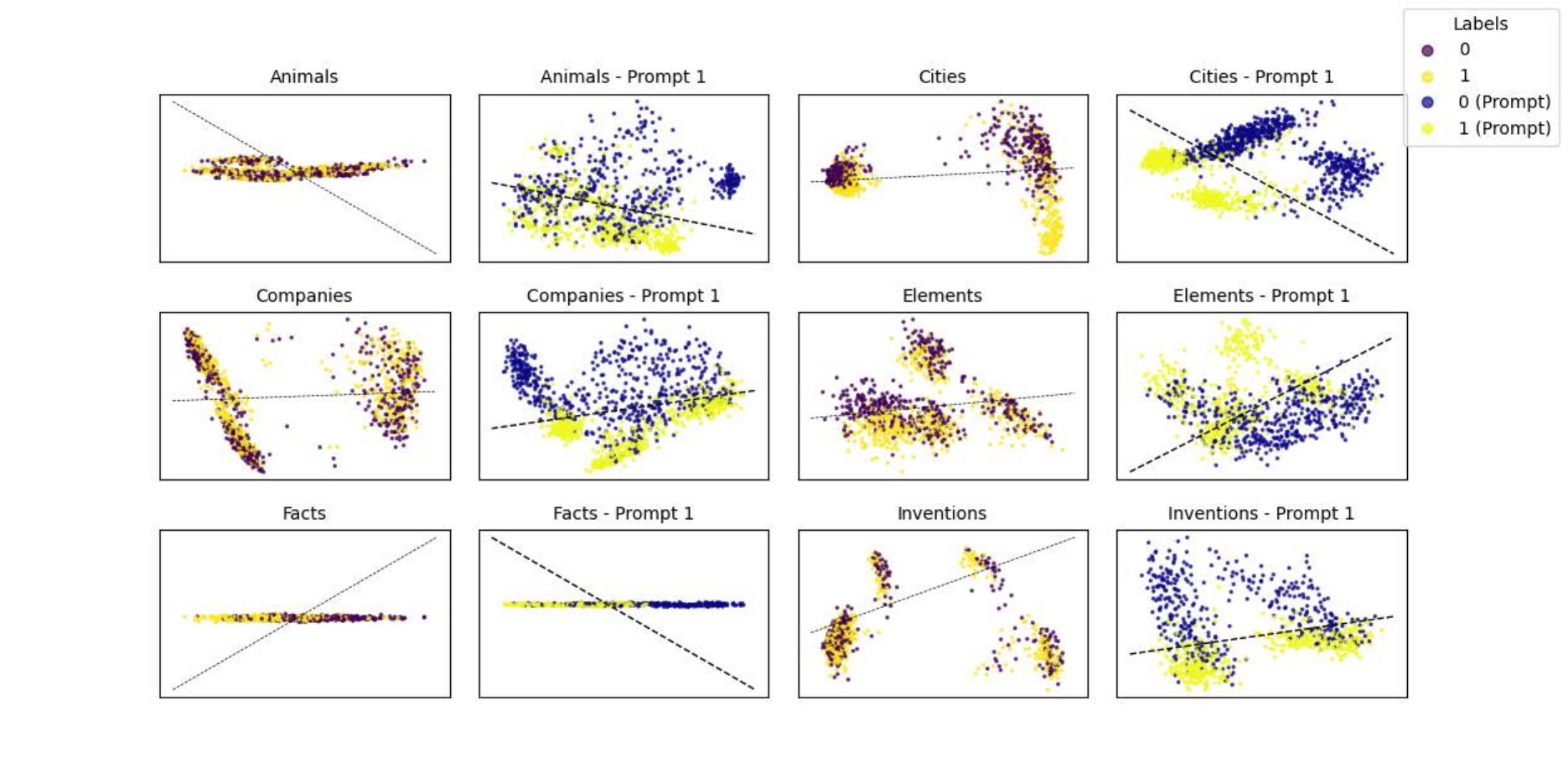}
    \caption{Visualization of the LLaMA2-7B-Chat model's internal states by using the 2-dimensional PCA on the True-False dataset before and after the introduction of Prompt 1, where label 0 represents false statements and label 1 represents true statements. Additionally, a logistic regression model was fitted to distinguish between true and false statements, with the decision boundary shown as a black dashed line.}
    \label{pic:PCA}
\end{figure*}

\section{Preliminary}
\label{sec:data}

In our research, to study the cross-domain hallucination detection problem, we utilized several publicly available datasets that encompass data from various domains. The first dataset we used is the True-False dataset~\cite{azaria2023internal}, which consists of six sub-datasets: "animals," "cities," "companies," "elements," "facts," and "inventions." These sub-datasets share similar textual structures but contain content on different topics. Each sub-dataset includes almost the same number of true and false statements, such as: 'Meta Platforms has headquarters in the United States' and 'Silver is used in catalytic converters and some electronic components.' More detailed information about this dataset can be found in Appendix~\ref{app:dataset}.

Another dataset we used is from~\citet{burger2024truth} and consists of 24 sub-datasets with 6 different topics: "animal\_class," "cities," "inventors," "element\_symb," "facts," and "sp\_en\_trans," as well as 4 different grammatical structures: affirmative statements, negated statements, logical conjunctions, and logical disjunctions. Affirmative statements were structured similarly to the examples in the True-False dataset, while negated statements were formed by negating the affirmative statements using the word "not." The sentences in logical conjunctions and logical disjunctions were constructed by sampling sentences from the affirmative statements and then connecting them with "and" or "or." The number of true and false statements within each grammatical structure is balanced. For clarity, we refer to this dataset as LogicStruct throughout this paper and present additional information about it in Appendix~\ref{app:dataset}.

%Due to data imbalance across topics, we focus more on grammatical structure variations within this dataset.
%Among the affirmative statements, "element\_symb," "animal\_class," "inventors," and "facts" are subsets of the datasets corresponding to these topics within the True-False datasets, while the other datasets were constructed based on the datasets used by~\citet{marks2023geometry} and~\citet{levinstein2024still}. 

% The sampling rules were carefully designed to ensure each dataset is balanced between true and false statements.

\section{Prompt-Guided Internal States}
\label{sec:prompt}

Several previous studies have focused on leveraging LLMs' internal states for hallucination detection~\cite{azaria2023internal,liu2024universal,DBLP:journals/corr/abs-2403-06448}. They assume that LLMs can recognize they have generated incorrect content, which is reflected in specific patterns in their internal states. Meanwhile, some studies have demonstrated the existence of an internal representation of truthfulness in LLMs~\cite{zou2023representation,marks2023geometry,burger2024truth}. Therefore, prior work typically selects contextualized embeddings corresponding to specific tokens in certain layers of LLMs, and then use them as features to train hallucination detectors. However, these embeddings were originally intended to guide text generation and they encode various information of the related text. As a result, the information related to text truthfulness becomes intertwined with domain-specific details, making it difficult for detectors to identify a consistent structure related to text truthfulness in LLMs' internal states.

Therefore, we hope to guide changes in LLMs' internal states such that the structure related to text truthfulness becomes more salient and consistent across texts from different domains. This would help detectors learn this structure and enhance their generalization performance. To achieve this goal, we experimented with a simple prompt:

\begin{small}
\begin{verbatim}
------------------------------------------------
Prompt 1:
Here is a statement: [statement]
Is the above statement correct?
------------------------------------------------
\end{verbatim}
\end{small}

\noindent This prompt directly asks LLMs the truthfulness of specific statements. When we input it into LLMs, the generation of new tokens will revolve around this question, which will be reflected in LLMs' internal states. 

%In this context, we hope that the contextualized embeddings corresponding to statements with different truthfulness exhibit more distinct pattern differences, leading to new structures related to text truthfulness in LLMs' internal states.

In the following two subsections, we will use various methods to demonstrate the changes in LLMs' internal states before and after the introduction of Prompt 1 and explain how these changes will effect the hallucination detection task.
Our analysis is conducted on the six sub-datasets of the True-False dataset. The feature vectors we used are the embeddings corresponding to the last token in the final layer of LLaMA2-7B-Chat.

\subsection{Effect of Prompt on Structural Salience}
\label{sec:pca}

%In this subsection, we focus on analyzing the salience of the structure related to text truthfulness in different datasets. We applied the technique of variance analysis to each dataset, revealing that the introduction of Prompt 1 can make this structure in LLMs' internal states more salient.

Some previous studies~\cite{marks2023geometry, burger2024truth} have shown that, on certain datasets, true and false statements have distinguishable geometric structures in LLMs' internal states and these structures can be directly observed after applying Principal Component Analysis (PCA) to reduce the dimensionality of relevant data. Therefore, we utilized the PCA to observe the salience of the structure related to text truthfulness before and after the introduction of Prompt 1.

From Figure~\ref{pic:PCA}, we can clearly observe that in all six datasets, the introduction of the prompt enables the first two principal components to effectively distinguish true and false statements. Since the first two principal components represent the directions with the largest variance among the embeddings, this indicates that the introduction of the prompt makes the structure related to text truthfulness in the embeddings more salient.
%However, this is just an intuitive but imprecise perspective, as we cannot be certain about the actual meaning represented by the first two principal components.

\begin{figure}[]
    \centering
    \resizebox{0.85\columnwidth}{!}{  
    \begin{tikzpicture}
        \begin{axis}[
            ybar,
            bar width=0.618cm,
            width=\textwidth,
            height=.618\textwidth,
            ylabel={Variance Ratio},
            ymin=0,
            ymax=0.3,
            symbolic x coords={Animals, Cities, Companies, Elements, Facts, Inventions, Average},
            xtick=data,  
            x tick label style={font=\LARGE, rotate=45, anchor=east},
            ylabel style={font=\LARGE},        
            legend style={font=\LARGE},        
            scaled ticks=false,
            y tick label style={/pgf/number format/fixed, /pgf/number format/precision=2}
            ]
            
            \addplot coordinates {
                (Animals, 0.0598)
                (Cities, 0.0928)
                (Companies, 0.0344)
                (Elements, 0.0650)
                (Facts, 0.0937)
                (Inventions, 0.0891)
                (Average, 0.0725)
            };
    
            \addplot coordinates {
                (Animals, 0.1396)
                (Cities, 0.2123)
                (Companies, 0.1360)
                (Elements, 0.1484)
                (Facts, 0.1771)
                (Inventions, 0.1227)
                (Average, 0.1560)
            };

            \legend{Before, After}
        \end{axis}
    \end{tikzpicture}
    }
    \caption{Comparison of the variance ratios before and after the introduction of Prompt 1 on each sub-dataset of the True-False dataset.}
    \label{fig:VR}
\end{figure}
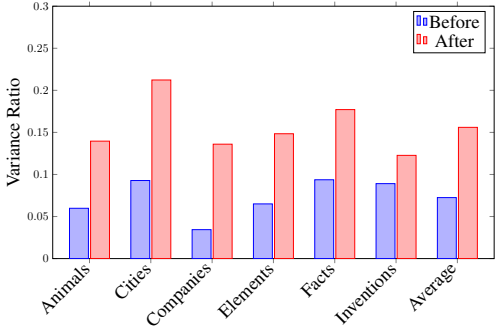

Moreover, to accurately describe the differences in structural salience, we performed a more detailed variance analysis. We believe that the proportion of the variance along a particular direction relative to the total variance can reflect the salience of the structure in that direction. Therefore, we subtracted the average embedding of false statements from the average embedding of true statements within each dataset, and referred to this direction as the 'truthfulness direction'. This was first used by~\citet{marks2023geometry} and its formula is as follows:

\begin{table*}[]
\caption{The cosine similarity between the truthfulness directions across different topics in the True-False dataset, before and after Prompt 1 addition.}
\label{tab:cos}
\centering
\setlength\tabcolsep{3pt} 
\fontsize{9pt}{9pt}\selectfont
\begin{tabular}{lcccccccccccc}
\toprule
                        & \multicolumn{6}{c}{Cosine Similarity Between Truthfulness Directions}                                                     & \multicolumn{6}{c}{Cosine Similarity After \textbf{Prompt 1} Addition}                                                       \\
                        \midrule
                        \textbf{Datasets}& \textbf{Animals} & \textbf{Cities} & \textbf{Comp.} & \textbf{Elements} & \textbf{Facts} & \multicolumn{1}{l|}{\textbf{Invent.}} & \textbf{Animals} & \textbf{Cities} & \textbf{Comp.} & \textbf{Elements} & \textbf{Facts} & \textbf{Invent.} \\
                        \toprule
\textbf{Animals}         & 1.0000 & 0.4368          & 0.4668           & 0.5498            & 0.6950           & \multicolumn{1}{l|}{0.3786 }          & 1.0000& 0.8037          & 0.7589           & 0.7284            & 0.8345           & 0.8333           \\
\textbf{Cities}         & 0.4368& 1.0000          & 0.4122           & 0.3300            & 0.4707           & \multicolumn{1}{l|}{0.2520}           & 0.8037& 1.0000          & 0.8349           & 0.7253            & 0.7274           & 0.8540           \\
\textbf{Comp.}         & 0.4668& 0.4122          & 1.0000           & 0.4302            & 0.5691           & \multicolumn{1}{l|}{0.4102}           & 0.7589& 0.8349          & 1.0000           & 0.7717            & 0.7255           & 0.8210           \\
\textbf{Elements}         & 0.5498& 0.3300          & 0.4302           & 1.0000            & 0.6258           & \multicolumn{1}{l|}{0.3301}           & 0.7284& 0.7253          & 0.7717           & 1.0000            & 0.8175           & 0.8071           \\
\textbf{Facts}           & 0.6950& 0.4707          & 0.5691           & 0.6258            & 1.0000           & \multicolumn{1}{l|}{0.4757}           & 0.8345& 0.7274          & 0.7255           & 0.8175            & 1.0000           & 0.7950           \\
\textbf{Invent.}           & 0.3786& 0.2520          & 0.4102           & 0.3301            & 0.4757           & \multicolumn{1}{l|}{1.0000}           & 0.8333& 0.8540          & 0.8210           & 0.8071            & 0.7950           & 1.0000           \\
\midrule
\textbf{Average}       & 0.5878& 0.4836          & 0.5481           & 0.5443            & 0.6394           & \multicolumn{1}{l|}{0.4744}           & \textbf{0.8265}& \textbf{0.8242}          & \textbf{0.8187}           & \textbf{0.8083}            & \textbf{0.8167}           & \textbf{0.8517}           \\
\toprule
\end{tabular}
\vspace{-5mm}

\end{table*}

\begin{equation}
\label{direction}
\theta = \frac{1}{N^+} \sum_{i=1}^{N^+} v_{i}^+ - \frac{1}{N^-} \sum_{i=1}^{N^-} v_{i}^-,
\end{equation}
where $\theta$ represents the truthfulness direction in the dataset $D$, $v_{i}^+$ and $v_{i}^-$ denote the embeddings corresponding to true and false statements, respectively, and $N^+$ and $N^-$ represent their counts.

Then we performed variance analysis along this direction. Let $v_i$ represent the embedding corresponding to a specific statement within dataset $D$, and let $N$ denote the total number of statements. By calculating
\begin{equation} 
\hat{v_i} = v_i - \frac{1}{N} \sum_{j=1}^{N} v_j ,
\end{equation}
we can use the de-centered vectors $\hat{v_i}$ to form a data matrix $X = (\hat{v_1}, \hat{v_2}, \dots, \hat{v_N})^T$, and then compute the covariance matrix corresponding to $X$:
\begin{equation}
\Sigma = \frac{1}{N-1} X^{T} X. 
\end{equation}
In this way, we obtain the total variance among all embeddings in dataset $D$:
\begin{equation} 
V_T = \text{Trace}( \Sigma ) 
\end{equation}
and the variance along the truthfulness direction:
\begin{equation} 
V_{\theta} = \frac{ \theta^T \Sigma \theta} {\|\theta\|^2} .
\end{equation}
Finally, we can use the ratio 
\begin{equation} 
\label{R}
R = \frac{ V_{\theta}} { V_T } 
\end{equation}
to represent the salience of the structure related to text truthfulness in dataset $D$. We calculated the corresponding variance ratios on the six sub-datasets of the True-False dataset and presented the results in Figure~\ref{fig:VR}.

We can observe that after the introduction of the prompt, this ratio shows a significant increase for all sub-datasets, indicating that the differences among the embeddings begin to concentrate more along the truthfulness direction. This enhances the salience of the structure related to text truthfulness in LLMs' internal states, making it easier for hallucination detectors to learn.

\subsection{Effect of Prompt on Structural Consistency}
\label{sec:cos}

%In this subsection, we focus on analyzing the consistency of the structure related to text truthfulness across different datasets. We calculated the cosine similarity between the truthfulness directions and observed that the introduction of the prompt makes these structures more consistent.

When we aim to apply detectors trained in one domain to others, the consistency of this structure becomes particularly important.
A more consistent structure will lead to better generalization performance. Therefore, we continued to utilize the 'truthfulness direction' defined by Equation (\ref{direction}) to analyze the consistency of this structure across different datasets and use the cosine similarity to represent this consistency:
\begin{equation}
c_{ij} = \frac{\theta_i \cdot \theta_j}{\|\theta_i\| \|\theta_j\|},
\end{equation}
where $c_{ij}$ represents the cosine of the angle between two truthfulness directions.

The calculation results for the six sub-datasets of the True-False dataset are presented in Table~\ref{tab:cos}. We can see that the introduction of the prompt significantly increases the cosine similarity between the truthfulness directions across different datasets, indicating that the structure related to text truthfulness between different datasets becomes more consistent. This consistency will help improve the generalization performance of related detectors, which use LLMs' internal states as input features to determine the truthfulness of statements.

\section{Methodology}

\label{sec:prism}

The analyses in sections~\ref{sec:prompt} indicate that the introduction of appropriate prompts can guide changes to the structure related to text truthfulness in LLMs' internal states, and these changes will facilitate our use of LLMs' internal states as features to perform the hallucination detection task. Based on this insight, we propose a novel framework \textbf{PRISM}, i.e., \underline{PR}ompt-guided \underline{I}nternal \underline{S}tates for hallucination detection of Large Language \underline{M}odels.

%This framework can effectively integrate the prompt-guided internal states with existing hallucination detection methods to enhance their generalization performance across data from different domains.

Our framework consists of two components:

First, we generate a large number of candidate prompt templates and select the one that best fits the current task. Specifically, we begin by manually constructing a prompt template $P$ related to the hallucination detection task, and then we use the language model $L$ to generate multiple candidate prompt templates $\{P_i\}_{i=1,\dots,N}$ that are similar in meaning but different in structure. For each $P_i$, we combine it with the text $a_j$ from the labeled dataset $A$ to obtain the text $P_i(a_j)$, and then input it into $L$ to obtain the corresponding feature vector $v_{ij}$ in the internal states. Subsequently, we use all feature vectors $\{v_{ij}\}_{j=1, \dots, M}$ to calculate the structure salience index $R_i$ corresponding to $P_i$ on $A$ and select the prompt template with the highest index as our final prompt template $P_s$.

Next, we integrate the prompt-guided internal states with existing hallucination detection methods to construct a more advanced detector. Specifically, based on the selected prompt template $P_s$, we extract feature vectors $\{v_{sj}\}_{j=1, \dots, M}$ corresponding to text $\{a_j\}_{j=1, \dots, M}$ in the dataset $A$ from the LLM's internal states. These feature vectors, together with their associated hallucination labels, are then used to train a hallucination detector $D$. For new text $T$ to be detected, we similarly combine it with the prompt template $P_s$ to obtain the text $P_s(T)$ and input it into the language model $L$ to obtain the corresponding feature vector $v$. Finally, we input $v$ into the hallucination detector $D$ to obtain its hallucination label $H$.

In the following subsections, we will provide a more detailed explanation of some key steps.

\begin{table*}[]
\caption{Average variance ratios for the 10 candidate prompt templates along with their ranking.}
\label{tab:prompt}
\centering
\setlength\tabcolsep{5.5pt} 
\fontsize{9pt}{9pt}\selectfont
\begin{tabular}{lccccccccccc}
\toprule
                        \textbf{Prompt}& \textbf{I} & \textbf{II} & \textbf{III} & \textbf{IV} & \textbf{V} & {\textbf{VI}} & \textbf{VII} & \textbf{VIII} & \textbf{IX} & \textbf{X} & \textbf{Without} \\

                        \midrule
\textbf{Ratio}         & 0.1093 & 0.1312         & 0.1291          & 0.1407            & 0.0875           & 0.1165           & 0.1041  & 0.1265 
 & 0.1228           & \textbf{0.1513}        & 0.0725               \\
\textbf{Ranking}         & 8 & 3         & 4          & 2            & 10           & 7           & 9  & 5 
 & 6           & \textbf{1}  & 11\\

\toprule
\end{tabular}
\vspace{-5mm}

\end{table*}

\subsection{Prompt Generation}

To obtain suitable candidate prompt templates for the hallucination detection task, we can use LLMs to assist in this process. First, we manually construct a simple prompt, such as Prompt 1 in Section~\ref{sec:prompt}. Next, we use this prompt to construct the following one:

\begin{small}
\begin{verbatim}
------------------------------------------------
Prompt 2:
"Here is a statement: '[statement]'
Is the above statement correct?"
This is a universal prompt template. Please
generate templates with similar meanings but 
diverse forms. The template should include the
embedding position of [statement].
------------------------------------------------
\end{verbatim}
\end{small}

\noindent Finally, we input the above prompt into LLMs to obtain a large number of prompt templates similar to Prompt 1. For example, we used Prompt 2 to query GPT-4o, resulting in 10 prompt templates that have similar meanings but different formats. The relevant content is provided in Appendix~\ref{app:prompt}.

\subsection{Prompt Selection}
\label{sec:select}

Once we have a large number of suitable prompt templates, we need to select the most appropriate one among them to perform our hallucination detection task.
A simple and effective method is to calculate the variance ratio corresponding to each prompt template on the labeled dataset using Equation (\ref{R}) and then leverage this ratio to guide our prompt selection. A higher ratio indicates that the prompt is more likely to achieve better results.

It is worth noting that this process relies on the selection of feature vectors from the internal states of LLMs. We selected the contextualized embedding corresponding to the last token in the final layer of LLaMA2-Chat-7B as the feature vector and calculated the corresponding variance ratio for each of the 10 prompt templates in Appendix~\ref{app:prompt}.
We computed the variance ratios for each prompt template across the six sub-datasets of the True-False dataset, averaged the results, and ranked them, as shown in Table~\ref{tab:prompt}.

According to Table~\ref{tab:prompt}, the 10th prompt template achieved the highest variance ratio and we present its form below:

\begin{small}
\begin{verbatim}
------------------------------------------------
Prompt 3:
Does the statement '[statement]' accurately 
reflect the truth?
------------------------------------------------
\end{verbatim}
\end{small}

\noindent We selected this prompt template for our framework in the experiments presented in Section~\ref{sec:exp}. We also conducted the same analysis on this prompt as we did in Section~\ref{sec:prompt} and presented the results in Appendix~\ref{app:p3}. 

\begin{table*}[]
    \centering
    \caption{The experimental results on the True-False dataset using LLaMA2-7B-Chat.}
    \setlength\tabcolsep{11pt} 
    \fontsize{9pt}{9pt}\selectfont 
    \begin{tabular}{lccccccc}
        \toprule
        \multirow{1}{*}{\textbf{Baselines}}
         & \textbf{Animals} & \textbf{Cities} & \textbf{Comp.} & \textbf{Elements} & \textbf{Facts} &\textbf{Invent.} & \textbf{Average} \\
        \midrule
        \multirow{1}{*}{\textbf{LN-PP}}
        & 0.5496 & 0.5653 & 0.5357 & 0.5727 & 0.5948 & 0.5527 & 0.5618 \\
        \multirow{1}{*}{\textbf{EUBHD}}
        & 0.5683 & 0.6197 & 0.5963 & 0.6004 & 0.5563 & 0.5591 & 0.5834 \\
        \multirow{1}{*}{\textbf{MIND}}
        & 0.4626 & 0.4664 & 0.4997 & 0.4850 & 0.5025 & 0.5023 & 0.4864 \\
        \multirow{1}{*}{\textbf{MM}}
        & 0.5300 & 0.5688 & 0.5648 & 0.5581 & 0.6137 & 0.5075 & 0.5572 \\
        \multirow{1}{*}{\textbf{SAPLMA}}
        & 0.6539 & 0.6700 & 0.6262 & 0.6057 & \textbf{0.7585} & 0.6233 & 0.6563 \\
        \midrule
        \multirow{1}{*}{\textbf{PRISM-MM}}
        & 0.7004 & \textbf{0.9060} & 0.7908 & 0.6385 & 0.7031 & 0.7756 & 0.7524 \\
        \multirow{1}{*}{\textbf{PRISM-SAPLMA}}
        & \textbf{0.7147} & 0.8936 & \textbf{0.8279} & \textbf{0.6705} & 0.7539 & \textbf{0.7804} & \textbf{0.7735} \\
        \bottomrule
    \end{tabular}
    \label{tab:results1}

\end{table*}

\begin{table*}[]
    \centering
    \caption{The experimental results on the True-False dataset using LLaMA2-13B-Chat.}
    \setlength\tabcolsep{11pt} 
    \fontsize{9pt}{9pt}\selectfont 
    \begin{tabular}{lccccccc}
        \toprule
        \multirow{1}{*}{\textbf{Baselines}}
         & \textbf{Animals} & \textbf{Cities} & \textbf{Comp.} & \textbf{Elements} & \textbf{Facts} &\textbf{Invent.} & \textbf{Average} \\
        \midrule
        \multirow{1}{*}{\textbf{LN-PP}}
        & 0.5538 & 0.5564 & 0.5397 & 0.5800 & 0.5954 & 0.5612 & 0.5644 \\
        \multirow{1}{*}{\textbf{EUBHD}}
        & 0.5684 & 0.5919 & 0.5835 & 0.5776 & 0.5589 & 0.5096 & 0.5650 \\
        \multirow{1}{*}{\textbf{MIND}}
        & 0.4957 & 0.4705 & 0.4372 & 0.5011 & 0.4986 & 0.4502 & 0.4756 \\
        \multirow{1}{*}{\textbf{MM}}
        & 0.5330 & 0.5366 & 0.5438 & 0.5327 & 0.6196 & 0.5085 & 0.5457 \\
        \multirow{1}{*}{\textbf{SAPLMA}}
        & 0.6584 & 0.7065 & 0.6721 & 0.6472 & \textbf{0.8054} & 0.6253 & 0.6858 \\
        \midrule
        \multirow{1}{*}{\textbf{PRISM-MM}}
        & 0.7111 & 0.8837 & \textbf{0.8498} & \textbf{0.7092} & 0.7804 & 0.7993 & 0.7889 \\
        \multirow{1}{*}{\textbf{PRISM-SAPLMA}}
        & \textbf{0.7405} & \textbf{0.8960} & 0.8435 & 0.6997 & 0.7918 & \textbf{0.8209} & \textbf{0.7987} \\
        \bottomrule
    \end{tabular}
    \label{tab:results2}

\end{table*}

\section{Experiment}
\label{sec:exp}

\subsection{Datasets and Metric}
We evaluate our PRISM and other baselines on the True-False and LogicStruct datasets which are detailed in section~\ref{sec:data}. Since the ratio of the two classes of labels in these datasets is close to 1:1, classification accuracy is an appropriate metric, which is calculated as:
\begin{equation}
\text{Acc} = \frac{N_r}{N},
\end{equation}
where $N$ denotes the total number of samples evaluated by the classifier, and $N_r$ denotes the number of correctly classified samples.

\subsection{Baselines}
We choose the following methods as baselines:
\begin{itemize}[leftmargin=*]

\item \textbf{LN-PP}~\cite{manakul2023selfcheckgpt} detects hallucinations based on the average probabilities of tokens generated by LLMs. We first determine a threshold, and then use it for label classification.

\item \textbf{EUBHD}~\cite{zhang2023enhancing} is a reference-free, uncertainty-based approach for detecting hallucinations in LLMs. This approach imitates human focus in factuality checking to estimate a hallucination score for the given text.

\item \textbf{MM}~\cite{marks2023geometry} uses a simple mass-mean probe to distinguish between true and false statements. This method selects specific activation values as feature vectors and uses Equation (\ref{direction}) to calculate the truthfulness direction on the labeled dataset. For the text to be detected, it first projects the feature vector onto the truthfulness direction by taking the dot product, then applies the sigmoid function, and finally rounds the result to obtain the label.

\item \textbf{SAPLMA}~\cite{azaria2023internal} is an effective supervised learning method. It uses activation values in LLMs as feature vectors to train a Multilayer Perceptron (MLP) network as the hallucination detector. %Selections of activation values from different layers can all achieve good classification performance.

\item \textbf{MIND}~\cite{DBLP:journals/corr/abs-2403-06448} is an unsupervised learning framework. It automatically generates labeled datasets from Wikipedia articles for training the hallucination detector. Other settings are consistent with the SAPLMA method.

\item \textbf{PRISM}. Our framework selects Prompt 3 from Section~\ref{sec:select} as the prompt template, and integrates with the SAPLMA and MM methods to obtain the PRISM-SAPLMA and PRISM-MM methods.

\end{itemize}

\subsection{Implementation Details}
\label{sec:implementation}
We conducted our experiments using LLaMA2-7B-Chat and LLaMA2-13B-Chat~\cite{touvron2023llama}. The activation values were selected from the contextualized embeddings corresponding to the last token in the final layer of the LLMs. In the SAPLMA and MIND methods, the dimensions of the detectors for each layer were 256, 128, 64, and 2, with a dropout rate of 20\% applied in the first layer. The activation function was ReLU, and the optimizer was Adam. These settings are the same as in the original paper~\cite{DBLP:journals/corr/abs-2403-06448}.

Our experiments focused on the generalization performance of different methods across different domains. For the SAPLMA method, we trained the detector separately on each sub-dataset of the True-False dataset and tested it on all other sub-datasets. Thus, for a single sub-dataset, we obtained multiple test results from detectors trained on other sub-datasets, and we averaged these results to obtain the final classification accuracy for this sub-dataset. In the LogicStruct dataset, we concentrated on training with affirmative statements and testing on other grammatical structures, as previous studies have indicated that achieving good generalization results in this scenario is challenging~\cite{marks2023geometry,burger2024truth}. The MM, LN-PP, and EUBHD methods adopted the same evaluation strategy as SAPLMA. Specifically, for the LN-PP and EUBHD methods, we determined an optimal threshold on the training set and applied it to the test set for classification. The MIND method was trained on its automatically generated dataset and tested on each sub-dataset of the True-False and LogicStruct datasets. The settings of the methods under PRISM framework remain consistent with those of the original methods.

For all detectors that required training, we split a validation set from the training set at a 4:1 ratio and selected the model parameters with the highest accuracy on the validation set over 10 training epochs as the test parameters. The final results presented are the averages obtained from training under three different random seeds.

\begin{table*}[]
    \centering
    \caption{The overall experimental results on the LogicStruct dataset.}
    \setlength\tabcolsep{9pt} 
    \fontsize{9pt}{9pt}\selectfont 
    \begin{tabular}{lcccc|cccc}  
    \toprule
    & \multicolumn{4}{c}{\textbf{LLaMA2-7B-Chat}} & \multicolumn{4}{c}{\textbf{LLaMA2-13B-Chat}} \\
    \midrule
    \textbf{Baselines}
     & \textbf{Neg.} & \textbf{Conj.} & \textbf{Disj.} & \textbf{Average} 
    & \textbf{Neg.} & \textbf{Conj.} & \textbf{Disj.} & \textbf{Average} \\
    \midrule
    \textbf{LN-PP}
       & 0.4108 & 0.5278 & \textbf{0.5743} & 0.5043 & 0.4263 & 0.5243 & 0.5863
 & 0.5123 \\
 \textbf{EUBHD}
       & 0.4184 & 0.5310 & 0.5387 & 0.4960 & 0.4512 & 0.5290 & 0.5263
 & 0.5022 \\
    \textbf{MIND}
       & 0.5219 & 0.4318 & 0.4969 & 0.4835 & 0.4849 & 0.4873 & 0.5038 & 0.4920 \\
    \textbf{MM}
       & 0.4948 & 0.4937 & 0.5013 & 0.4966 & 0.5071 & 0.4895 & 0.5020 & 0.4995 \\
    \textbf{SAPLMA}
       & 0.4864 & 0.5196 & 0.5030 & 0.5030 & 0.5155 & 0.6788 & 0.5186 & 0.5710 \\
    \midrule
    \textbf{PRISM-MM}
       & 0.5548 & \textbf{0.8087} &0.5580 & 0.6405 & 0.5260 & \textbf{0.8734} & \textbf{0.6207} & 0.6734 \\
    \textbf{PRISM-SAPLMA}
       & \textbf{0.7345} & 0.7529 & 0.5329 & \textbf{0.6734} & \textbf{0.7389} & 0.8112 & 0.5722 & \textbf{0.7074} \\
    \bottomrule
    \end{tabular}

    \label{tab:results3}
\end{table*}

\begin{table*}[]

    \caption{The experimental results of all 10 prompt templates on the True-False dataset using LLaMA2-7B-Chat.}
    \centering
\setlength\tabcolsep{5.3pt} 
\fontsize{9pt}{9pt}\selectfont
\begin{tabular}{lccccccccccc}
\toprule
                        \textbf{Prompt}& \textbf{I} & \textbf{II} & \textbf{III} & \textbf{IV} & \textbf{V} & {\textbf{VI}} & \textbf{VII} & \textbf{VIII} & \textbf{IX} & \textbf{X} & \textbf{Without} \\

                        \midrule
\textbf{Acc.}         & 0.7217 & 0.7678         & 0.7297          & 0.7398            & 0.6997           & 0.7749           & 0.6931  & 0.7162 
 & 0.7003           & 0.7524        & 0.5572               \\
\textbf{Ranking}         & 6 & 2         & 5          & 4            & 9           & 1           & 10  & 7 
 & 8           & 3  & 11\\

\toprule
\end{tabular}
\vspace{-5mm}
\label{tab:prompt-all}

\end{table*}

\subsection{Experimental Results}
\label{sec:results}
We can see from Table~\ref{tab:results1} and~\ref{tab:results2} that our framework significantly outperforms other baselines in terms of the classification accuracy on the True-False dataset. 
\textbf{On almost every sub-dataset of different topics (5 out of 6 topics), our framework substantially improves the generalization performance of the original methods.} Additionally, there is a significant difference in the performance of the original MM and SAPLMA methods, which may be due to the simpler structure of the detector used in the MM method compared to that in the SAPLMA method. However, after integrating both into our framework, these two methods achieve similar performance. According to Section~\ref{sec:pca}, this may be because the introduction of the prompt makes the structure related to text truthfulness more salient in the LLMs' internal states, making it easier for detectors to capture this structure.
On the Facts sub-dataset, the original SAPLMA method and our framework achieve nearly identical accuracy. As shown in Figure~\ref{pic:PCA}, before the introduction of prompts, the true and false statements in the Facts sub-dataset already exhibit clear separability, so it is difficult to further enhance this structure by appropriate prompts.

Table~\ref{tab:results3} presents the experimental results on the LogicStruct dataset. We can observe that, except for our framework, other baselines struggle to generalize the training results on affirmative statements to other grammatical structures. According to the observations made by~\citet{burger2024truth}, this is because the structure related to text truthfulness in affirmative statements is different from the corresponding structure in other grammatical structures.
However, \textbf{our framework achieves significant generalization performance across different grammatical structures.} This indicates that the introduction of the prompt indeed makes the structure related to text truthfulness more consistent across different datasets, which aligns with the observation in Section~\ref{sec:cos}.
%Additionally, we can observe that the average results on the 13B model outperform those on the 7B model, suggesting that as the model's reasoning capabilities improve, our framework is likely to perform better as well.

\subsection{Ablation Studies}

\subsubsection{Impact of Prompt Selection }
In this section, we focus on the impact of prompt selection. We conducted experiments using the PRISM-MM method on all 10 prompt templates presented in Appendix~\ref{app:prompt}, with the results presented in Table~\ref{tab:prompt-all}.
We can observe that all the prompt templates significantly improved the average accuracy compared to the result of the original MM method. This indicates that the impact of prompts on improving the generalization performance of hallucination detectors is general and stable.

In addition, by combining the ranking information from Table~\ref{tab:prompt} and Table~\ref{tab:prompt-all}, we can see that the top 4 prompt templates selected based on the variance ratios all perform within the top 5 in actual experiments, while the two worst-performing prompt templates correspond to the two lowest variance ratios. Furthermore, we calculated the Pearson correlation coefficient between the variance ratios in Table~\ref{tab:prompt} and the actual experimental accuracies reported in Table~\ref{tab:prompt-all}. The resulting correlation coefficient is 0.7708, with a statistically significant p-value of 0.0055. These results indicate that the variance ratio can help us select prompt templates that achieve better generalization results in actual experiments while avoiding poorly performing ones.

\subsubsection{Impact of Layer Selection }

In addition to selecting the contextualized embeddings corresponding to the last token in the final layer of LLMs as feature vectors, we also extracted the embeddings corresponding to the last token in the middle layer (16th) of LLaMA2-7B-Chat to evaluate our framework. We conducted the same experiments on the True-False dataset with our framework and the corresponding original methods,  as shown in the last four rows of Table~\ref{tab:abla}.

Compared to the results in Table~\ref{tab:results1}, the performance of the original methods has worsened, while our framework has achieved better results. This indicates that our framework exhibits good stability with respect to different layer selections.

\subsubsection{Impact of Internal States}
After introducing the prompt, is it possible to assess the truthfulness of the text directly based on LLMs' responses without relying on their internal states? To investigate this question, after inputting the prompt into LLaMA2-7B-Chat, we directly extracted the token probabilities for " Yes" and " No" when the LLM was going to generate the next token and calculated their ratio \( p[3869]/p[1939] \), where $p$ represents the vocabulary probabilities for generating the next token. We used whether the ratio is greater than 1 to determine the label of the original text. The average accuracy on the True-False dataset is shown in the first row of Table~\ref{tab:abla}.

We can observe a significant accuracy difference between directly using the responses generated by the LLM and training detectors based on the prompt-guided internal states. This indicates that the LLM's internal states contain more information related to text truthfulness, which is not observable from the LLM's responses but still plays a crucial role in hallucination detection.

\begin{table}[t]
    \centering
    \caption{The experimental results on the True-False datase using feature vectors from the middle layer of LLaMA2-7B-Chat and the next token probabilities from LLaMA2-7B-Chat.}
    \label{tab:abla}
    \setlength\tabcolsep{8pt} 
    \fontsize{9pt}{9pt}\selectfont  
    \begin{tabular}{lcc}
        \toprule
        \multirow{1}{*}{\textbf{Baselines}}
        & \textbf{Average Accuracy} \\
        \midrule
        \multirow{1}{*}{\textbf{Yes/No}}
        & 0.6166 \\
        \midrule
        \multirow{1}{*}{\textbf{MM(middle)}}
        & 0.5166 \\
        \multirow{1}{*}{\textbf{SAPLMA(middle)}}
        & 0.5766 \\
        
        \midrule
        \multirow{1}{*}{\textbf{PRISM-MM(middle)}}
        & 0.7924 \\
        \multirow{1}{*}{\textbf{PRISM-SAPLMA(middle)}}
        & \textbf{0.7938} \\
        \bottomrule
    \end{tabular}
     
\end{table}

\subsubsection{Impact of Real-world Scenarios}
 
To evaluate the effectiveness of our framework in real-world scenarios, we conducted experiments on the TruthfulQA dataset~\cite{lin-etal-2022-truthfulqa}, comparing the performance of the MM, SAPLMA, PRISM-MM, PRISM-SAPLMA, and SelfCheckGPT methods. The SelfCheckGPT method is an effective unsupervised method that detects hallucinations by assessing the consistency among multiple responses generated by the LLM for the same question. However, since the formats of texts used in previous experiments were not in a question-answer format, this method could not be applied there.

The TruthfulQA dataset consists of 790 question-answer pairs drawn from real-world scenarios, such as: "Why do veins appear blue?" The questions span 37 diverse categories that are particularly prone to hallucinated responses, including Misconceptions and Superstitions. We used the LLaMA2-7B-Chat model with greedy sampling to generate answers for each question and labeled them based on BLEURT scores by comparing the generated responses to the reference answers, following the original paper. We set the BLEURT classification thresholds to 0, 0.25, and 0.5, respectively. Responses with scores above the threshold were labeled as correct, while those below were considered hallucinations.

For the SelfCheckGPT method, we followed the original paper’s setup by generating 20 additional samples at temperature T=1 for each question and used GPT-4 to assess the consistency among responses and compute a hallucination score for the target response. For the MM and SAMPLA methods, we trained the detector on the first 20\% of the data and tested it on the remaining 80\%. Since the question types are ordered sequentially, this setup effectively reflects the detector’s ability to generalize across domains. The PRISM framework adopted Prompt 3 from Section~\ref{sec:select} as the prompt template. To ensure robustness, the results were averaged over three random seeds after training for 10 epochs. Due to label imbalance in the dataset, we used AUROC as the evaluation metric. The results are presented in Table~\ref{tab:real}.

\begin{table}[]
    \centering
    \caption{The experimental results (AUROC) on the TruthfulQA dataset using LLaMA2-7B-Chat with different BLEURT thresholds.}
    \label{tab:real}
    \setlength\tabcolsep{8pt} 
    \fontsize{9pt}{9pt}\selectfont  
    \begin{tabular}{lccc}
        \toprule
        \multirow{1}{*}{\textbf{Baselines}}
        & \textbf{T = 0} & \textbf{T = 0.25} & \textbf{T = 0.5}\\
        \midrule
        \multirow{1}{*}{\textbf{SelfcheckGPT}}
        & 0.5937 & 0.6168 & 0.6357 \\

        \multirow{1}{*}{\textbf{MM}}
        & 0.5654 & 0.5559 & 0.5279 \\
        \multirow{1}{*}{\textbf{SAPLMA}}
        & 0.6389 & 0.6603 & 0.6654 \\
        
        \midrule
        \multirow{1}{*}{\textbf{PRISM-MM}}
        & 0.6808 & \textbf{0.7082} & 0.7114 \\
        \multirow{1}{*}{\textbf{PRISM-SAPLMA}}
        & \textbf{0.6887} & 0.6939 & \textbf{0.7248} \\
        \bottomrule
    \end{tabular}
     
\end{table}

As we can see, on this dataset, the PRISM framework significantly improves the generalization performance of the original hallucination detectors, and outperforms the SelfCheckGPT method. This demonstrates that our framework retains its effectiveness even in more complex, real-world hallucination scenarios.

\section{Conclusions}

In this paper, we first investigated the effect of specific prompts on the structure related to text truthfulness in LLMs' internal states. We found that suitable prompts can make this structure more salient and consistent across datasets from different domains, facilitating hallucination detectors in learning this structure and improving their generalization performance.
Based on this insight, we introduced a novel framework, prompt-guided internal states for hallucination detection of large language models, namely PRISM.
This framework integrates the prompt-guided internal states with existing hallucination detection methods to obtain more advanced detectors.
Finally, we conducted experiments across various baselines on datasets from different domains, and the results demonstrated that our framework can significantly enhance the generalization performance of existing hallucination detection methods. %We hope that our framework can inspire future research in LLMs hallucination detection.

\section*{Limitations}

We acknowledge that our research still has some limitations. 
When detecting hallucinations in the content generated by LLMs, we heavily rely on their internal states, which may be challenging to access in some cases, requiring us to use a proxy model. Additionally, our framework does not leverage other information produced by LLMs during text generation, such as probability information and the generated text itself. In future research, we will explore fully utilizing various types of information produced by LLMs during text generation to achieve more effective hallucination detection.

\section*{Ethics Statement}
The development and application of the PRISM framework for Large Language Models hallucination detection are guided by the principle of assessing the reliability and credibility of AI-generated content. PRISM aims to reduce the risk of users being misled by AI-generated content, thereby contributing to safer and more trustworthy AI applications. Additionally, the datasets used in this project are publicly available and do not involve personal or sensitive data, ensuring privacy and security. 
We need to recognize that our framework classifies text based on the model's internal states, and thus may inherit biases from the language model on certain issues, potentially leading to incorrect judgments. Therefore, we encourage further research and collaboration to develop ethical, fair, and trustworthy AI systems.

% Bibliography entries for the entire Anthology, followed by custom entries
%\bibliography{anthology,custom}
% Custom bibliography entries only
%\bibliography{custom}

\clearpage

\appendix

\section{Dataset Details}
\label{app:dataset}

The following are some examples of statements from the True-False dataset:
\begin{itemize}
    \item Animals: "The gazelle has distinctive orange and black stripes and is an apex predator."
    \item Cities: "Oranjestad is a city in Aruba."
    \item Companies: "Meta Platforms has headquarters in United States."
    \item Elements: "Silver is used in catalytic converters and some electronic components."
    \item Facts: "The planet Jupiter has many moons."
    \item Inventions: "Narinder Singh Kapany invented the programming language (theoretical)."
\end{itemize}
We can observe that the displayed statements contain both true and false information, and the relevant details are shown in  Table~\ref{tab:true}.

\begin{table}[t]
\caption{ \small{The information of the True-False dataset, where T=0 indicates that the sentence is annotated as false, and T=1 indicates true.}}
\label{tab:true}
\centering
\footnotesize{\begin{tabular}{cccc}
\toprule
\textbf{Topic}  & \textbf{\#Sentence} & \textbf{\#T=0} &\textbf{\#T=1}  \\
                   \midrule
\text{Animals}                & 1008    & 504 & 504                            \\
\text{Cities}                     & 1458  & 729 &729                              \\
\text{Companies}      & 1200      & 600  & 600                           \\
\text{Elements}         & 930    & 465  & 465                             \\
\text{Facts}         & 613   & 307  &306            \\
\text{Inventions}     & 876     & 412  & 464                           \\

\toprule
\end{tabular}}
\end{table}

Examples of sentences with different grammatical structures in the LogicStruct dataset are as follows:
\begin{itemize}
    \item Affirmative statements: "The Spanish word ’dos’ means ’enemy’."
    \item Negated statements: "The Spanish word ’dos’ does not mean ’enemy’."
    \item Logical conjunctions: It is the case both that [statement 1] and that [statement 2].
     \item Logical disjunctions: It is the case either that [statement 1] or that [statement 2].
\end{itemize}
In both Logical conjunctions and Logical disjunctions, statement 1 and statement 2 are derived from Affirmative statements and sampled in a way that ensures label balance across each dataset. Additional information is presented in Table~\ref{tab:neg}.

\begin{table}[t]
\caption{ \small{The information of the LogicStruct dataset, where \{Topic\} in the Name refers to "animal\_class," "cities," "inventors," "element\_symb," "facts," and "sp\_en\_trans," and \#Sentence represents the total number of data across all topics for each grammatical structure.}}
\label{tab:neg}
\centering
\footnotesize{\begin{tabular}{ccc}
\toprule
\textbf{Name}  & \textbf{Description} & \textbf{\#Sentence}   \\
                   \midrule
\text{\{Topic\}}                &  \text{Affirmative statements}    & 3167                             \\
\text{\{Topic\}\_neg}                & \text{Negated statements}    & 3167                            \\
\text{\{Topic\}\_conj}                     & \text{Logical conjunctions}  & 3998                            \\
\text{\{Topic\}\_disj}      & \text{Logical disjunctions}      & 3000                          \\

\toprule
\end{tabular}}
\end{table}

\section{Analysis of Prompt 3}
\label{app:p3}
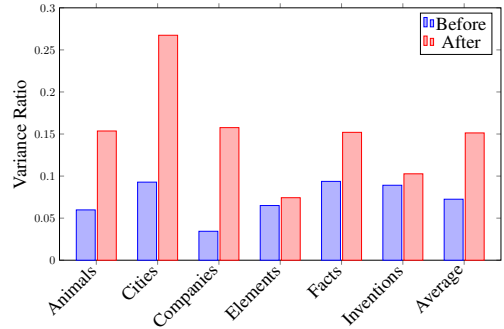
\begin{figure}[b]
    \centering
    \resizebox{0.85\columnwidth}{!}{  
    \begin{tikzpicture}
        \begin{axis}[
            ybar,
            bar width=0.618cm,
            width=\textwidth,
            height=.618\textwidth,
            ylabel={Variance Ratio},
            ymin=0,
            ymax=0.3,
            symbolic x coords={Animals, Cities, Companies, Elements, Facts, Inventions, Average},
            xtick=data,  
            x tick label style={font=\LARGE, rotate=45, anchor=east},
            ylabel style={font=\LARGE},        
            legend style={font=\LARGE},        
            scaled ticks=false,
            y tick label style={/pgf/number format/fixed, /pgf/number format/precision=2}
            ]
            
            \addplot coordinates {
                (Animals, 0.0598)
                (Cities, 0.0928)
                (Companies, 0.0344)
                (Elements, 0.0650)
                (Facts, 0.0937)
                (Inventions, 0.0891)
                (Average, 0.0725)
            };
    
            \addplot coordinates {
                (Animals, 0.1536)
                (Cities, 0.2674)
                (Companies, 0.1576)
                (Elements, 0.0743)
                (Facts, 0.1520)
                (Inventions, 0.1027)
                (Average, 0.1513)
            };

            \legend{Before, After}
        \end{axis}
    \end{tikzpicture}
    }
    \caption{Comparison of the variance ratios before and after the introduction of Prompt 3 on each sub-dataset of the True-False dataset.}
    \label{fig:VR3}
\end{figure}

Consistent with the analysis of Prompt 1 in Section~\ref{sec:prompt}, we calculated the variance ratios of Prompt 3 on each sub-dataset of the True-False dataset. The results before and after the introduction of Prompt 3 are presented in Figure~\ref{fig:VR3}. We also calculated the cosine similarity between the truthfulness directions across these sub-datasets after the introduction of Prompt 3, as shown in Table~\ref{tab:cos3}.

We can observe that Prompt 3 significantly enhances the salience and consistency of the structure related to text truthfulness in the embeddings. This is consistent with the conclusion in Section~\ref{sec:prompt}, indicating that using prompts to guide changes in LLMs' internal states is generally effective.

\begin{table}[t]
\caption{The cosine similarity between the truthfulness directions across different topics in the True-False dataset after Prompt 3 addition.}
\label{tab:cos3}
\centering
\setlength\tabcolsep{1pt} 
\fontsize{9pt}{9pt}\selectfont
\begin{tabular}{lcccccc}
\toprule

                        \textbf{Datasets}& \textbf{Animals} & \textbf{Cities} & \textbf{Comp.} & \textbf{Elements} & \textbf{Facts} & \textbf{Invent.} \\
                        \toprule
\textbf{Animals}     &    1.0000& 0.8370          & 0.8043           & 0.8144            & 0.8519           & 0.8811           \\
\textbf{Cities}           & 0.8370& 1.0000          & 0.8113           & 0.7835            & 0.8570           & 0.8613           \\
\textbf{Comp.}            & 0.8043 & 0.8113          & 1.0000           & 0.8012            & 0.7927           & 0.8192           \\
\textbf{Elements}           & 0.8144 & 0.7835          & 0.8012           & 1.0000            & 0.8327           & 0.8154           \\
\textbf{Facts}           & 0.8519 & 0.8570          & 0.7927           & 0.8327            & 1.0000           & 0.8522           \\
\textbf{Invent.}          & 0.8811 & 0.8613          & 0.8192           & 0.8154            & 0.8522           & 1.0000           \\
\midrule
\textbf{Average}         & \textbf{0.8265}& \textbf{0.8242}          & \textbf{0.8187}           & \textbf{0.8083}            & \textbf{0.8167}           & \textbf{0.8517}           \\
\toprule
\end{tabular}
\vspace{-5mm}
\end{table}

\clearpage

\section{Generation of Prompt Templates}
\label{app:prompt}
%The complete conversation with GPT-4o during the process of generating suitable prompt templates is shown at https://chatgpt.com/share/672279f0-3528-8005-8695-cb6d72d0c713. Before starting the conversation, we ensured that GPT's long-term memory was cleared. The detailed dialogue is as follows:

\begin{minipage}{0.95\textwidth}
\noindent\fbox{
\parbox{\textwidth}{
\textbf{Question:} \\"Here is a statement: '[statement]' \\
Is the above statement correct?" \\
This is a universal prompt template. Please generate templates with similar meanings but diverse forms. The template should include the embedding position of [statement]. \\
\textbf{GPT-4o:} \\Here are various templates with diverse forms that ask about the correctness or validity of a given statement:

1. "Is it accurate to say that '[statement]'?" \\
2. "Would you consider the statement '[statement]' to be correct?"\\
3. "Can we confirm that '[statement]' is true?"\\
4. "Does the statement '[statement]' hold true?"\\
5. "Is '[statement]' a valid statement?"\\
6. "Is there accuracy in the claim '[statement]'?"\\
7. "Could '[statement]' be considered a factual statement?"\\
8. "Is it correct to assume that '[statement]' is true?"\\
9. "Would it be right to say '[statement]' is accurate?"\\
10. "Does the statement '[statement]' accurately reflect the truth?"\\
Each version maintains the focus on verifying the truth or accuracy of the statement in question.
}
}
\end{minipage}

\end{document}